\documentclass[sigconf]{acmart}

\usepackage{color}
\definecolor{ForestGreen}{rgb}{0.13, 0.55, 0.13}
\newcommand{\newgd}[1]{{{#1}}}

\usepackage{soul,color}
\setstcolor{red}
\setul{}{.2ex}

\AtBeginDocument{%
  \providecommand\BibTeX{{%
    \normalfont B\kern-0.5em{\scshape i\kern-0.25em b}\kern-0.8em\TeX}}}

\setcopyright{acmcopyright}
\copyrightyear{2023}
\acmYear{2023}
\acmDOI{XXXXXXX.XXXXXXX}

\acmBooktitle{Communications of the ACM}
\acmPrice{15.00}
\acmISBN{978-1-4503-XXXX-X/22/06}

\begin{document}

\title{\newgd{Data} Bias Management}
\author{Gianluca Demartini}
\affiliation{%
  \institution{The University of Queensland}
  \country{Australia}
}
\email{demartini@acm.org}


\author{Kevin Roitero}
\affiliation{%
  \institution{University of Udine}
  \country{Italy}
}
\email{kevin.roitero@uniud.it}

\author{Stefano Mizzaro}
\affiliation{%
  \institution{University of Udine}
  \country{Italy}
}
\email{mizzaro@uniud.it}






\begin{abstract}
Due to the widespread use of data-powered systems in our everyday lives, concepts like  bias and fairness gained significant attention among researchers and practitioners, in both industry and academia. Such issues typically emerge from the data, which comes with varying levels of quality, used to train supervised machine learning systems.
With the commercialization and deployment of such systems that are sometimes delegated to make life-changing decisions, significant efforts are being made towards the identification and removal of possible sources of data bias that may resurface to the final end user or in the decisions being made.
In this paper, we present research results that show how bias in data affects end users, where bias is originated, and provide a viewpoint about what we should do about it. We argue that data bias is not something that should necessarily be removed in all cases, and that research attention should instead shift from bias removal towards the identification, measurement, indexing, surfacing, and adapting for bias, which we name \emph{bias management}.
\end{abstract}





\maketitle

\section{Introduction}
The presence of bias in data has led to a lot of research being conducted to understand the impact of bias on machine learning models and data-driven decision making systems \cite{baeza2018bias}. 
Research has focused on questions related to the fairness in the decisions taken by models trained with biased data, and on designing methods to increase the transparency of automated decision making processes so that possible  bias issues  may be easily spotted and ``fixed'' by removing bias.

\newgd{Recent approaches taken in the literature to deal with data bias first aim to understand the cause of the problem (e.g., a sub-set of the population being under-represented in the training dataset) and then propose and evaluate an ad-hoc intervention to reduce or remove the bias from the system (e.g., by selecting which additional training data items to label in order to re-balance the dataset and increase equality (i.e., a balanced representation of classes) rather then equity (i.e., over-representing the disadvantaged sub-set of the population).}

Example research on bias removal include work looking at 
how to remove bias from learned word embeddings. \citet{bolukbasi2016man} defined a methodology for modifying an embedding representation to remove
gender stereotypes. This allows researchers to remove certain bias (e.g., the association between the words `receptionist' and
`female') while preserving other gender-relation information (e.g., maintaining the associating between the words `queen' and `female'). 
\citet{sutton2018biased} observed how gender bias which is often present in professions is also reflected in word embeddings and proposed to remove gender bias from embeddings by using a projection function. 
\citet{schick2021self} proposed an algorithm that reduces the probability of a language model generating problematic text.
These are examples of methods that enable researchers and practitioners to intervene, subjectively, by making individual choices on how and where bias should be removed from models created from bias-reinforcing human-generated datasets.

\newgd{Additional examples include the removal of bias from  prediction datasets where researchers decided how to re-balance datasets to increase fairness across groups when doing data augmentation \cite{pastaltzidis2022data}, feature augmentation \cite{fong2021fairness}, or adjusting the metrics that measure bias \cite{lum2022biasing}. All these approaches include personal choices made by the researchers on how to detect and what to do with bias.}

As compared to these examples of the body of work that has looked at bias removal, in this paper we aim at proposing a different perspective on the problem by introducing the task of \textit{bias management} rather than that of removal. The rationale lies in the fact that bias is inevitably present in human annotations (as we show in Section \ref{sec:bias}).
Thus, making a deliberate choice to remove bias deploying certain interventions or post-processing collected labels in a certain way, would introduce a bias itself as there may be multiple ways in which biased labels may be curated.
Instead, in Section \ref{sec:management}, we propose an alternative approach to manage bias in data-driven pipelines, that increases data and bias transparency thus allowing us to surface it to end users and human decision-makers empowering them to take reparative actions and interventions themselves by leveraging common-sense and contextual information. This brings away the responsibility from the data workers and system developers that risk to bias the results according to their perspective instead.

\section{An Example: Bias in Search}\label{sec:search}
Consider the case where, to create an image dataset, a  data scientist needs to collect manual annotations or label a set of images; it is reasonable to assume that such a dataset may then be used to train an automatic system to independently perform a specific object identification task in fresh, unseen images.
Suppose then that a user issues the gender-neutral query ``nurse'' to an image search engine trained on this dataset. The user may see on the search engine result page the vast majority of images being of female nurses. 
While this might appear as an indication that the ranking algorithm of the search engine has a gender bias issue, as it would be qualified by the vast majority of existing bias/fairness measures \cite{AMIGO2023103115}, this might also reflect the real gender distribution of people employed in this profession, that is, for example, female nurses are statistically more frequent than male nurses. 
While a traditional approach might look at resolving this bias by forcing the algorithm to show a balanced result set with male and female nurses in a similar percentage, we argue that an alternative, less invasive algorithmic approach might be more useful to the end users.

The search engine might display on the result page a set of additional metadata which may be useful to the user to have a complete understanding of the magnitude of bias in the search result set; for example, the search engine might show a label indicating that ``the search results appear to be highly imbalanced in terms of gender: in the top 1,000 results, 870 of them are of female nurses and 130 of them are of male nurses; \newgd{this is however similar to the gender distribution in the nursing profession where official data from your country government shows that 89\% of the workforce in 2016 was female.}''. This information makes the end user more informed and aware of the statistical distribution of the search results with respect to a specific group (in this case, gender) also increasing gender bias literacy \newgd{and the understanding of current societal norms}. Then, ideally, the user should be asked by the system if they would like to maintain the current search result or whether they would prefer to inspect the results after a fairness policy of their choice is applied to the data (in this case, for example, forcing the number of male and female nurses to be roughly the same in the search result list). 
These, or similar, remarks can be transferred from search systems to recommender systems.

Research has looked at bias in search and at how gender-neutral queries may return gender-imbalanced results \cite{otterbacher2017competent}. More than that, \citet{otterbacher2018investigating} looked at how end users of search engines perceive biased search results. They studied how users perceive gender-imbalanced search results as compared to how they score on a scale of sexism. Their results show how different people perceive the results differently, thus confirming that a one-size-fits-all bias removal solution would not be appropriate for all users. A similar approach would apply to content in social media feeds.

We argue that not employing an explicit and transparent bias removal intervention might even be potentially harmful to the user. In fact, if the task of the end user in our example scenario was to investigate something related to or influenced by the percentage of male and female nurses, the implicit application of the fairness policy as decided by the system designer might leave the user with an inaccurate perception of the real gender distribution in the nursing profession. Taking this concept to the extreme, the user might even erroneously think, somehow paradoxically, that gender bias is not present in the nursing profession, and that male and female nurses are equally present on this job market. \newgd{This triggers ethical questions related to how we should manage bias, which we discuss in Section \ref{sec:ethics}}.

A related study by \cite{silberzahn2018many} looked at how groups of data analysts reach different conclusions when working on the same dataset and trying to answer a given question (i.e., `Are soccer referees more likely to give red cards to dark skin toned players than light skin toned players?' in this specific piece of research). They found that different groups made very different observations which led the authors to conclude that when analyzing complex data it may not be possible to avoid reaching diverse conclusions. This example supports the need for something different from a one-size-fits-all approach to bias.

\newgd{\section{A Cause of Unfairness: \\ Bias in Human Annotations}}\label{sec:bias}
Next, we look at some examples of bias in human annotations. This serves the purpose to explain how human annotations, which are often used to train and evaluate the performance of machine learning models, carry bias and stereotypes from the human annotators providing the labels.

Crowdsourcing is a popular way to collect input from human annotators and contributors. A common example of a successful crowdsourcing project is Wikipedia and its related projects.
\newgd{Wikipedia is known to have a gender-biased population of editors where the majority are men\footnote{\url{https://en.wikipedia.org/wiki/Gender_bias_on_Wikipedia}}.}
As a research example, \citet{sarasua2019evolution} studied participation bias in Wikidata showing not only that participation is very skewed with very few editors contributing most of the content and a long-tail of very many editors that contribute little, but also that the way contributions are made by these two different groups (i.e., head and tail of the distribution) varies substantially.


Such long-tailed distribution of participation in crowdsourcing project has shown to be common also in paid micro-task crowdsourcing \cite{franklin2011crowddb} which is often used to collect labels to train supervised machine learning models.
This means that very few human annotators end up contributing the majority of the labels in the dataset.
\newgd{This is a common behavior pattern also known as Nielsen’s 90-9-1 participation rule
\footnote{\url{https://www.nngroup.com/articles/participation-inequality/}} stating that in this type of projects 90\% of users often are observers, 9\% are sporadic contributors, and 1\% account for most of the contributions.}
We  discuss in Section 4 how this may be problematic from a data bias point of view.

Thus, the data stored in crowdsourced knowledge graphs like Wikidata is influenced by the (imbalanced) population of contributors.
This then leads to data imbalance in terms of class representation. \citet{luggen2019non} has shown how to estimate class completeness in Wikidata and observed how certain classes may be more complete than others.

A similar approach may be used to measure the gender balance of entities in a knowledge graph like Wikidata (e.g., how many female astronauts are there in the dataset?). Once the measurements is done, intervention actions may be taken. For example, Wikidata editors may decide to only contribute new entities for the class Astronauts that are female until a balance is achieved between genders. Alternatively, editors may focus on having equal completeness rates for all genders. (e.g., 80\%  of all female astronauts and 80\% of male astronauts rather than having a higher completion rate for one and lower for the other gender). While measurements can be done algorithmically, intervention decisions are human-made and who makes the adjustment decisions is another source of bias that is to be reflected in the underlying data.

Training supervised machine learning model with this kind of unbalanced training data is a typical cause of Unknown Unknowns (UUs) errors. 
These are errors made with high model confidence, thus showing how the model is unaware of the possibility of making classification mistakes. UUs often appear for classes which are poorly represented in the training data and may lead to fairness issues, where certain segments of the population under-represented in the training data systematically gets more wrong decisions (e.g., a minority population in loan decisions).
While humans can be deployed to identify UUs \cite{han2021iterative}, the common fix to the problem is the collection of further labels or the augmentation of existing training data for the under-represented classes.

\begin{figure}
    \centering
    \includegraphics[width=0.45\textwidth]{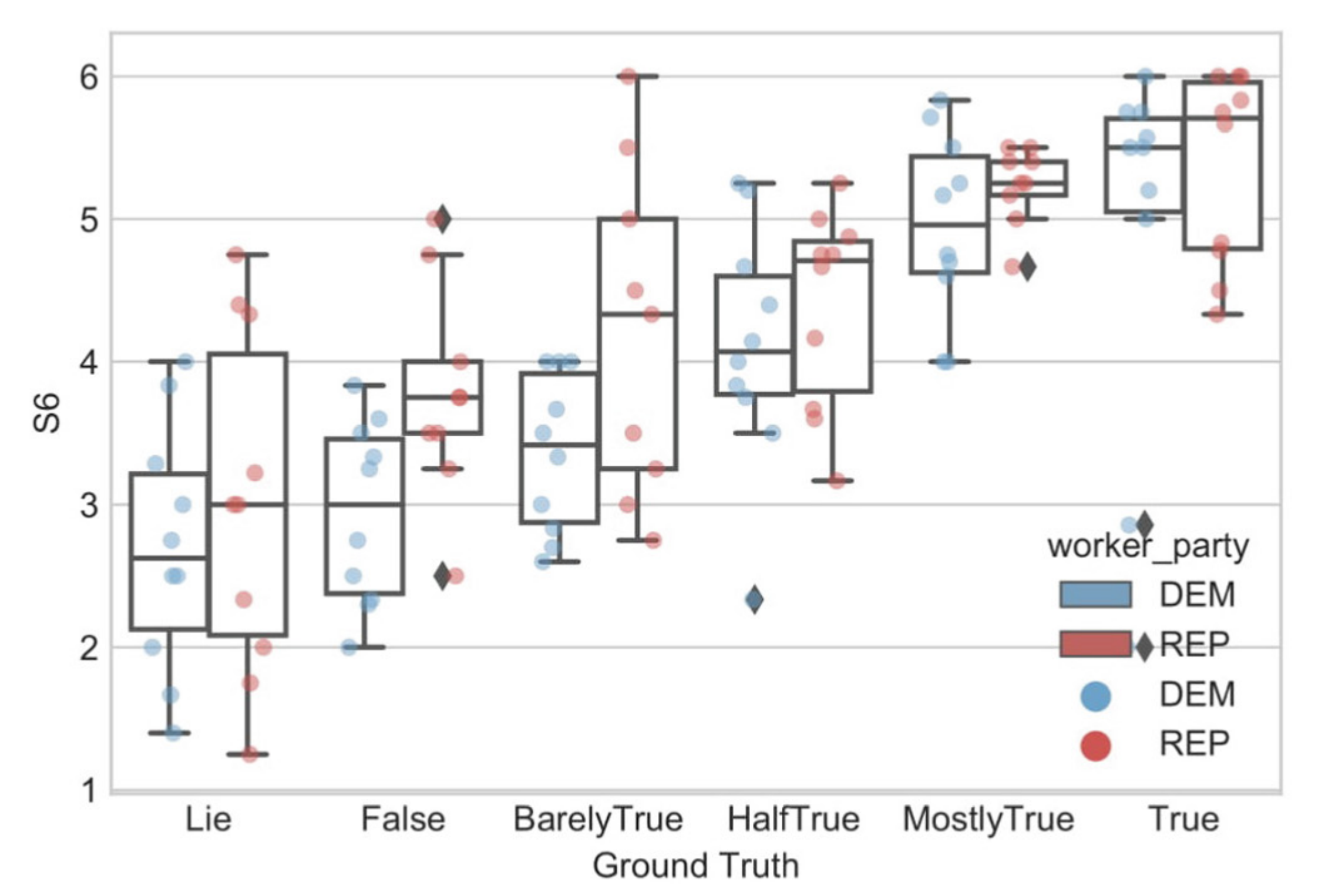}
    \caption{Crowdsourced misinformation labels by non-experts (y-axis) as compared to expert fact-checkers (x-axis) for statements made by Republican politicians (from \cite{barbera2020crowdsourcing}).}
    \label{fig:politicalbias}
\end{figure}

\begin{figure*}
    \centering
    \includegraphics[width=0.9\textwidth]{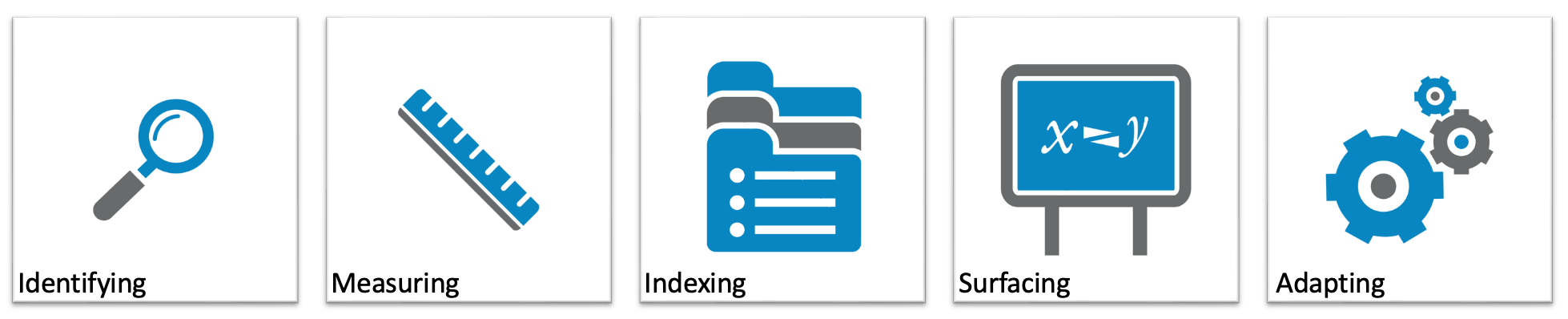}
    \caption{The five steps of bias management.}
    \label{fig:steps}
\end{figure*}

\newgd{\section{A Source of Biases: \\ Human Annotators}}
\label{sec:annotators}
Vast amounts of labelled data are necessary to train large supervised machine learning models.
This labelled data typically comes from human annotators.
Relevant research questions are related to the way human annotators annotate data, how different annotators annotate differently, and how different-than-usual data gets annotated.

\citet{barbera2020crowdsourcing} looked at how non-expert human annotators label misinformation and observed a political bias. That is, unsurprisingly, people perceive misinformation differently based on their political background. As shown in Figure~\ref{fig:politicalbias}, human annotators that vote Republican are more generous when labelling the level of truthfulness of statements by Republican politicians (and the same is true the other way around). We can see that, systematically at all levels of truthfulness, the scores assigned by the Republican crowd (in red) are higher than those given by annotators with a Democrats background (in blue).
This example shows not only how different annotators provide different labels, but also the presence of systematic bias in human annotators.

Another relevant study by \citet{fan2022socio} has looked at how `unusual' data gets manually annotated. They made use of short videos depicting people washing their hands in developing and developed countries setting up a study controlling for the socio-economic status of the people depicted in the videos. Then, they collected labels from humans annotators based in the US and observed the presence of bias in the labels. They observed, for example, how  videos showing people in Africa receive more negative annotations than those from Asia, how videos with higher-income families receive more positive annotations, and how high-income households received more descriptive annotations.
This example shows how data may receive different annotations that reflect bias and stereotypes present in human annotators.
\newgd{These examples of annotation bias are often the result of cognitive biases (confirmation bias in particular) \cite{nickerson1998confirmation}.}

A way to track the source of bias is by means of collecting data about the annotation process and the involved annotators. Recent research has made use of logs of human annotation behaviour data to study the annotation process.
For example,
\citet{han2020modelling} looked at editing behaviour in Wikidata,  and \citet{han2020understanding} at how data scientists curate data. These two studies consistently showed that the results are different based on who the individual person providing labels and making decisions is.

These studies confirm the conclusion that different human annotators would provide diverse labels for the same dataset, and that models trained on such labels would provide different decisions \cite{perikleous2022does}. Thus, selecting the right mix of human annotators can lead to less biased labels and, as a result, less biased automated decision-support systems.

\section{How to Deal with Bias}\label{sec:management}

Rather than removing biased information or avoiding to do so, we believe it is a better option to keep track of bias and surface it to the end users. This would serve the purpose of increasing transparency over the entire data pipeline, rather than having an algorithmic fix to the bias problem.
%
The example of `bias in search' presented in Section \ref{sec:search} shows how interventions may have consequences on end users and decision-makers and, the interventions themselves, may introduce bias.

Thus, our proposal to deal with bias and produce resilient data pipelines \cite{sadiq2022information} is based on the assumption that the algorithmic results should not be subjectively changed by potentially biased design choices,  but rather enriched with metadata that can surface information about bias to the end users and information consumers. In this way, they would be empowered to make their own choices and interventions on the system.

\subsection{A Bias Management Pipeline}
Our proposal, alternative to removal of bias, consists of five different step (see Figure \ref{fig:steps}. 

\begin{enumerate}
    \item \emph{Identifying}: identify if the data or system being used is subject to bias or fairness issues. Where does the data come from? Who is providing the annotations? 
    \item \emph{Measuring}: quantify with an appropriate metric the magnitude of different types of bias present in the data or system which is under consideration. How does the label distribution look like across classes and annotators?
    \item \emph{Indexing}: collect, parse, structure, and store bias metadata and fairness policies aimed at facilitating a subsequent fast and effective retrieval and system adaptation. For each piece of data, who has labelled it? What are their background attributes? How can the bias be adapted for or resolved?
    \item \emph{Surfacing}: present in an appropriate way to the end user the bias present in the underlying data and/or any fairness policy that have been applied to the  data or system  under consideration. How many items from one class are there in the top-k results?
    \item \emph{\newgd{Adapting}}: provide the user with a set of tools which allows them to interact with existing biased results and to adapt them for bias in their preferred ways. This enables them to make  informed decisions. Giving control to the user is essential since for some tasks they may benefit from fairness (e.g., a job application scenario) while for some others they may not (e.g., understanding the gender distribution in a specific profession). Does the user want to see a balanced distribution or a representative distribution of the results?
\end{enumerate}

\newgd{Recent research has started to look at some of these steps already. For example, on the front of measuring bias, \citet{lum2022biasing} have shown how current measures that try to quantify model performance differences for different parts of a population are themselves statistically biased estimators and new ways to measure bias are needed.}

\newgd{\subsection{The Ethics of Adapting for Bias}}\label{sec:ethics}
\newgd{One important question that has to be raised is about the appropriateness of surfacing bias metadata to end users. In certain situations this may lead to negative behaviour and potentially to harm. For example, not all people may be comfortable in being exposed to evidence of discrimination existing in the society, or in the dataset they are looking at, and could instead feel safer when presented with culturally-aligned data, even if that potentially reinforces stereotypes.}

\newgd{When dealing with bias present in data and results, deciding about the most appropriate way to adapt for it is a culturally-dependent, subjective decision and should be the result of each individual's preference. For example, when adapting for gender biased search results, we might want to consider that women in STEM are the majority in many Iranian and Indian universities, but this may not be the case in other locations.}
\newgd{Similar questions should be asked by system designers when applying personalization techniques. For example, \citet{reinecke2011improving} have shown in a controlled study how users were 22\% faster when using a culturally adapted user interface.}

\newgd{In the end, the goal would be to empower the end users and to provide them with informative, and potentially perspective-changing adaptive strategies, based on their own preferences on which adjustments should be applied or not to any existing data bias.}
\newgd{To deal with these challenges we claim that such adaptations should be an individual user choice made available to them by the data-driven systems, but finally chosen by the end users according to their own preference, believes, and comfort.}

\section{Conclusions}
We argue that bias is part of human nature, and that it should be managed rather than removed as removal would instead introduce a different type of bias by the system designers and engineers making ad-hoc choices \cite{johansen2020studying}.
The bias management model detailed in this paper envisions a different approach from the current bias and fairness research.  We believe that the ideas detailed in this paper can lead to a more sound, informed, and transparent data-driven decision making process which will impact future data pipeline design.


\bibliographystyle{ACM-Reference-Format}
\bibliography{sample-base}

\balance

\end{document}